# MODULATED BINARY CLIQUENET

*Jinpeng Xia* [1, 4], *Jiasong Wu* [1, 2, 3, 4], *Youyong Kong* [1, 4], *Pinzheng Zhang* [1, 4], *Lotfi Senhadji* [2, 3, 4], *Huazhong Shu* [1, 4]

[1]LIST, the Key Laboratory of Computer Network and Information Integration (Southeast University), Ministry of Education, 210096 Nanjing, China
[2]INSERM, U1099, Rennes, F-35000, France
[3]Université de Rennes 1, LTSI, Rennes, F-35042, France
[4]Centre de Recherche en Information Médicale Sino-français (CRIBs), Southeast University, Université de Rennes 1, INSERM

## ABSTRACT

Although Convolutional Neural Networks (CNNs) achieve effectiveness in various computer vision tasks, the significant requirement of storage of such networks hinders the deployment on computationally limited devices. In this paper, we propose a new compact and portable deep learning network named Modulated Binary Cliquenet (MBCliqueNet) aiming to improve the portability of CNNs based on binarized filters while achieving comparable performance with the full-precision CNNs like Resnet. In MBCliqueNet, we introduce a novel modulated operation to approximate the unbinarized filters and gives an initialization method to speed up its convergence. We reduce the extra parameters caused by modulated operation with parameters sharing. As a result, the proposed MBCliqueNet can reduce the required storage space of convolutional filters by a factor of at least 32, in contrast to the full-precision model, and achieve better performance than other state-of-the-art binarized models. More importantly, our model compares even better with some full-precision models like Resnet on the dataset we used.

## 1. INTRODUCTION

Convolutional neural networks (CNNs) have been widely used in the computer vision field to deal with various tasks because of their powerful ability to represent features from pixels. Behind the powerful representation capabilities, a large amount of data is under support. CNNs usually contain over millions of parameters and hundreds of Megabyte storage space. This poses great difficulties for deploying neural networks on embedded platforms.

Many effective models like Alexnet [1], Resnet [2], Inception [3] and Densenet [4] have been proposed over the past few years. Cliquenet [5], inspired by Densenet, extracts more refined features, because of the architecture with not only forward-intensive connections but also reverse-intensive connections to refine previous levels of information. In the same Clique block [5], there are forward and reverse connections between any two layers, which improves the flow of information in the deep network.

Binary weights instead of real-value weights have been investigated in CNNs to compress models [6]-[15]. For examples, Courbariaux et al. [6] introduced BinaryConnect, which involves training deep neural networks (DNNs) with binary weights during forward and backward propagation while maintaining the accuracy of the storage weights of the cumulative gradients. Based on BianryConnect, Binary Neural Network (BNN) [7] was proposed to train a CNN with binary weights and activations at inference-time and at train-time full-precision parameters are used to compute gradient. Further, Xnor-net [9] is an architecture where both the input and weights of the convolution are constrained to binary values and approximates the unbinarized filters with a single scaling factor. McDonnell [14] proposed 1-bit Wide Residual Networks (WRN) in which the standard deviation changes from $\sqrt{2/(w^2 I)}$ to 1 after binary process and then scaled the output of convolutional layers with a constant rather than a single scaling factor as an improvement of Xnor-net [9]. Wang et al. [15] proposed Modulated Convolutional Networks (MCNs) which gives a similar concept of M-filters and uses a matrix to approximate the unbinarized filters instead of a single scaling factor in Xnor-net [9].

Aiming to reduce the model storage space via binarized filters and improve the performance of the compressed model, in this paper, we propose a new binary architecture named Modualted Binary Cliquenet (MBCliqueNet) to compress the neural networks based on parameter sharing, binarized convolutional filters and M-filters. The differences between MBCliqueNet and Cliquenet are listed as follows: 1) We replace the last fully-connected layer in Cliquenet with a $1 \times 1$ convolutional layer and a batch normalization layer, which further improve the performance of Cliquenet. 2) We replace the convolutional parameters with sign function and use 1-bit weights so that the proposed MBCliqueNet reduces the storage space of a full-precision model. 3) We define a modulate process based on a special M-filter to approximate the full-precision convolutional filters with binarized weights, which also leads to the reduction of convolutional parameters. 4) We add only one M-filter in one block shared with all convolutional layers in the block to reduce the extra parameters. 5) We propose an initialization method of M-filters to speed up the converge of the model and prove that our model can be learned by the backpropagation algorithm. As a result, the proposed MBCliqueNet reduces the storage space of a full-precision model by a factor of 32 and reduce the parameters due to the special modulate progress, while achieving the state-of-the-art performance on MNIST, SVHN, Cifar10 and Cifar100 datasets in terms of classification accuracy, compared to the existing binarized CNNs and even the full-precision models like Resnet.

The rest of the paper is organized as follows. The architecture of the proposed MBCliqueNet is described in Section 2. The performance of MBCliqueNet is analyzed with respect to various size of M-filters and also compared to other architectures in Section 3. Section 4 concludes the paper.

## 2. PROPOSED METHODS

The proposed MBCliqueNet with three blocks have been illustrated in Fig. 1. The architecture is mostly based on Cliquenet [5] but we replace the fully connected layer with a $1 \times 1$ convolutional layer and a batch normalization layer. The transition layers include a convolution and an average pooling to change the sizes of feature maps. In each block there are 4 layers and their convolutional filters are binarized by sign function, and then take a modulate process with a shared M-filter to reconstruct convolutional filters which are used on feature maps. The output feature maps of each block are Stage-II feature as stated in Cliquenet [5] and shown in Fig. 2. With the multi-scale strategy, we take features from 3 blocks after a global pooling operation and then fuse them with a concatenation operation. Different from Cliquenet [5], we send the fused features to a $1 \times 1$ convolution layer and a batch normalization layer instead of a fully-connected layer.

## 2.1. Modulated Binary Cliqueblock

In Clique Block [5], any two layers in the block have bidirectional connection except for the input node. Compared with Clique Block, the convolutional filters are binarized and we add a M-filter in the Modulated Binary Cliqueblock to modulate all the layers in the block. The propagation of a Modulated Cliqueblock with 3 layers is shown in Fig. 1. The output $X_i$ of the $i$th ($i \geq 1$) layer in the $s$th ($s \geq 1$) stage can be formulated as:

$$X_i = \begin{cases} g\left(\sum_{l<i}(M \circ \widehat{W}_{li}) \oplus X_l\right), s = 1 \\ g\left(\sum_{l<i}(M \circ \widehat{W}_{li}) \oplus X_l^k + \sum_{m>i}(M \circ \widehat{W}_{mi}) \oplus X_m^{k-1}\right), s \geq 2 \end{cases} \quad (1)$$

where ∘ represents the modulated operation and ⊕ represents the convolutional operation. $g$ is the non-linear activation function. $M$ is the M-filter in the block and $\widehat{W}_{li}$ is the binarized convolutional parameters between the $l$th layer and the $i$th layer ($i \geq 1$).

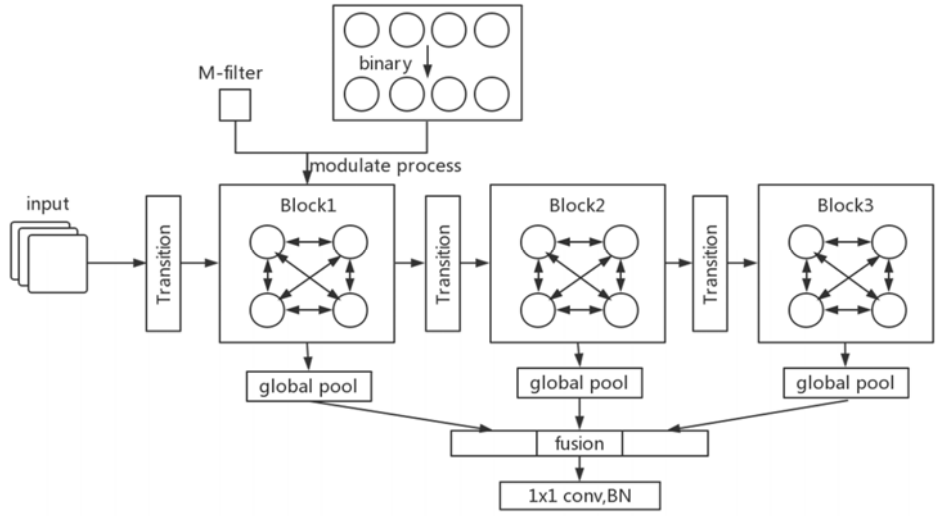

Fig. 1. MBCliqueNet with three blocks

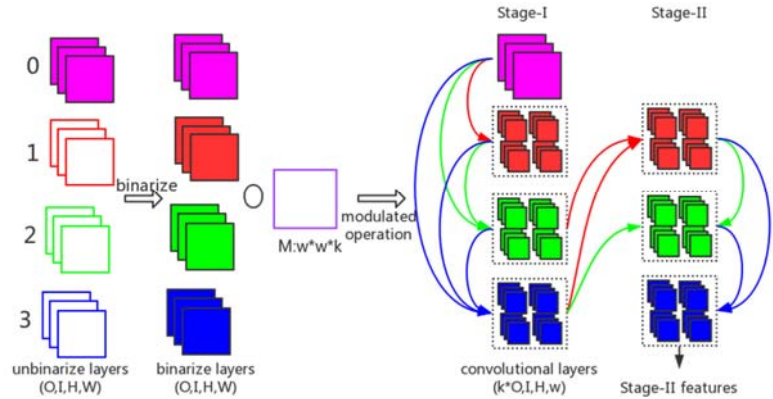

Fig. 2. Modulated binary block with 3 layers. The node 0 represents the input feature maps of the block and no operation is conducted on it. The convolutional filters are binarized and make a modulated operation with M-filters to reconstruct filters which are used for convolution products. Nodes 1, 2 and 3 are in a recurrent connect, both input and output of each other. The modulated operation will increase the number of filters from $O$ to $kO$, where $O$ represents the original number of filters and $k$ represents the channel of the M-filter.

Table 1. Classification accuracy (%) on 4 datasets

| Method | MBCliquenet | Binaryconnect [6] | BNN [7] | Cliquenet [5] | MCNs [15] | ResNet101 [2] | 1-bit WRN [14] |
|---|---|---|---|---|---|---|---|
| Params | 5.2M | - | - | 6.94M | 12.7M | - | 26.8M |
| SVHN | 97.21 | 97.85 | 97.49 | 98.44 | 96.87 | - | 98.07 |
| MNIST | 99.63 | 98.99 | 98.60 | - | 99.52 | - | - |
| Cifar10 | 95.72(95.78) | 91.73 | 89.85 | 94.90 | 95.39 | 93.57 | 95.28 |
| Cifar100 | 80.43(80.69) | - | - | 76.68 | 78.13 | 74.85 | 80.65 |

The Modulated operation is used to reconstruct the convolutional filters. Following [9], we use the M-filter matrix to be the weight of the binarized filters instead of the α used in [9]. The ∘ operation is defined as:

$$M \circ \widehat{W} = (Q_1, \dots, Q_{kO}) \quad (2)$$

$M$ represents a M-filter we designed with the size of $k \times w \times w$, where $k$ represents the number of channels and each channel is a $w \times w$-sized 2D filter. $\widehat{W}$ represents the binarized convolutional filters with the size of $O \times I \times w \times w$, where $I$ is the channel of the input features and $O$ represents the number of filters. $Q$ represents the reconstructed filters. By doing this operation one binarized convolutional filters, we use a M-filter with $k$ channels to reconstruct $k$ convolutional filters. So the operation $M \circ \widehat{W}$ finally reconstruct $kO$ filters. The reconstructed filter $Q$ is defined as:

$$Q_{ki+j} = M'_j * \widehat{W}_i \quad (3)$$

where index $i$ represents the filter index with $i = 0, \dots, O-1$. $M'_j$ represents the $I$ copies of the 2D matrix $M_j$ with $j = 1, \dots, k$. ∗ represents the element-wise multiplication operator (Schur product operation). With the same number of output feature maps, $1/k$ parameters will be used.

Compared with [15], our modulated process is based on one filter with size of $I \times w \times w$ rather than the fixed size $k \times w \times w$. This leads to a different convolutional product that is more efficient in terms of calculations.

The output feature maps $X^{l+1}$ of the $l$th layer are calculated by the reconstructed filters $Q^l$ in the $l$th layer as:

$$X_k^{l+1} = Q_i^l \oplus X^l \quad (4)$$
$$X^{l+1} = [X_1^{l+1}, \dots, X_{kO}^{l+1}] \quad (5)$$

where $X_k^{l+1}$ represents the $k$th channel of $X^{l+1}$ and $Q_i^l$ represents the $i$th reconstructed filter in the $l$th layer.

### 2.2. Binary method and parameter initialization

We use the sign function [9] to binary the full-precision weights. However, when using the sign function alone, there is a problem in controlling gradient and activation scaling through the network [14], especially when we add a modulated matrix.

To solve the problem, we consider the initialization method of M-filters and full-precision convolutional filters. Following [14], we make full-precision convolutional filters obeying a Gaussian distribution with the mean value of 0 and the standard deviation of $\sqrt{2/(w^2 I)}$. After applying the sign function on full-precision filters, the binarized filters have as mean 0 and as standard deviation 1. To make the reconstructed filters having the same standard deviation as the full-precision filters, we initialize the M-filter in the block with the mean of 0 and the standard deviation of $\sqrt{2/(w^2 I_{block})}$, where $I_{block}$ represents the unified channel numbers of the full-precision filters in this block. We add a $1 \times 1$ convolutional kernel between each layer to change the channels of input features so that we can set the channel numbers of each layer in this block with the same number. Through this initialization method, we enable the standard deviation of forward-propagation features in the binarized network to be equal to that we would have in full-precision network.

Note that because of the M-filter added in each block and the specific initialization method, we have to solve the problem with the increasing channels of the input features. Hence we adopt the $1 \times 1$ convolution before the $3 \times 3$ convolution which has the same input channels as the other layers in the block. Before the convolution operation, there are batch normalization and Relu layers as usual.

### 2.3. Backpropagation

Following [6], [7] and [9], we find out that there will be good results when we apply sign function to full-precision weights to get 1-bit-per-weight during training but update the full-precision at the backpropagation time. So in our MBCliqueNet, we only update the unbinarized filters first and then M-filters.

We define $L_S$ to be the loss function and $\delta_W$, $\delta_M$ to be the gradient of the unbinarized filters $W_i$ and the corresponding M-filter.

$$\delta_W = \frac{\partial L_S}{\partial W_i} = \frac{\partial L_S}{\partial Q} \cdot \frac{\partial Q}{\partial W_i} = \sum_j \frac{\partial L_S}{\partial Q_{ij}} \cdot M_j \quad (6)$$

where $Q_{ij}$ is the $j$th corresponding reconstructed filters by $W_i$ and $M'_j$ represent the $I$ copies of the 2D matrix $M_j$ with $j = 1, \dots, k$.

We further calculate the gradient of the M-filter with fixed $W_i$,

$$\delta_M = \frac{\partial L_S}{\partial M} = \frac{\partial L_S}{\partial Q} \cdot \frac{\partial Q}{\partial M} = \sum_{i,j} \frac{\partial L_S}{\partial Q_{ij}} \cdot W_i \quad (7)$$

where $W_i$ represents the $i$th filter in the block and $Q_{ij}$ represents the $j$th reconstructed filter correspond to $W_i$

Eqs. (6) and (7) show the proposed MBCliqueNet can be learned with the backpropagation algorithm.

### 3. EXPERIMENTAL RESULTS

#### 3.1. Experimental datasets and training details

We conducted experiments on five datasets: MNIST [16], SVHN [17], Cifar10 [18], Cifar100 [18], and Tiny ImageNet [19].

We implement our models in the Pytorch platform and training it on two 1080Ti GPUs. We train our models with the standard stochastic gradient descent methods (SGD) used in [20]. We use cross-entropy as our loss function, mini-batch size of 128, momentum of 0.9 and weight decay of 0.0005. For all datasets, we use simple data augmentation insisting of randomly flipping each training data with probability 0.5, and padding each image on all sides by 4 pixels and crop images from a random location.

We use a 'warm restart' learning rate schedule reported in [20] that allows to speeding up the convergence. The schedule periodically changes the learning rate from the maximum learning rate to the minimum one according to a cosine decay, across a certain number of epochs, and then restore the learning rate to the maximum to repeat the process across many epochs. We set the learning rate to vary from 0.1 to 0.0001 and set period to reduce the learning rate to be 2 epochs, and then 4, 8, 16, 32, 64, 128 epochs. For MNIST and SVHN datasets, we only train on 32 epochs and for Cifar10/100 we train on 256 epochs.

### 3.2. Effect of M-filters' channel $k$

The MBCliqueNet is based on the modulated operation which will reduce the parameters of convolutional filters to $1/k$. So the choice of $k$ will influence the performance of the model. We conduct experiment on the same model with different choices of $k$ on the Cifar10 dataset. For fair, we use 3 blocks, 4 layers in each block and set the widths of each block to be 64-64-128. The width represents the channels of output feature maps of each convolutional layers in each block so that the truly number of convolutional filters is $1/k$ as the width. The number of parameters used (params) and the accuracy of results are reported in Table 2.

Table 2. Results of different $k$

| $k$ | 2 | 4 | 8 |
|---|---|---|---|
| Params | 1.6 M | 1.3 M | 1.1 M |
| Accuracy (%) | 95.17 | 95.03 | 94.89 |

With the increase of $k$ we figure out that the accuracy decays in an acceptable range by 0.15% but the reduction of parameters is considerable. Due to the existence of fully-connected layers, the parameters are not $k$ times reduced. Note that even for $k = 8$, our model is still better than the full-precision Resnet-101 (93.57%) [2] with similar amount of parameter.

### 3.3. Results on MNIST and SVHN

For MNIST and SVHN, the size of M-filters is $4 \times 3 \times 3$ ($k = 4$). The MBCliqueNet has 3 blocks containing each 4 layers with widths of 64-64-128. The input image is through a transition layer to change the channels of the input image, which contains a convolutional layer with kernel sizes of $3 \times 3$, stride of 1, padding of 1 and widths of 64. The width is set to be the same as the first block.

Note that the width here represents the output channels of feature maps. Due to the specific modulated operation we designed, the truly width of the convolutional filters is $64/k$. The results are shown in Table 1. On the MNIST dataset, our model achieves the best result compared with other models. On the SVHN dataset, it performances are better than MCNs [15] with 0.4% improvement with less parameters. Furthermore, our model is only trained on 32 epochs but the MCNs [15] is trained on 200 epochs.

### 3.4. Results on Cifar10 and Cifar100

For Cifar10 and Cifar100 datasets, we use more parameters to explore the best performances of our model. The architecture of MBCliqueNet has 3 blocks, 4 layers in each block, widths of each block are 64 – 128 - 256. We set the channels of M-filter $k$ to be 4. We change the minimum of accuracy to $1 \times 10^{-5}$. The accuracy of MBCliqueNet achieves 95.72% and 80.43% on Cifar10 and Cifar100, respectively, as shown in Table 1. The training and testing curves are shown in Fig. 3. From Table 1, we find out that our MBCliqueNet achieves the best performance in contrast to the other state-of-the-art binary methods both on Cifar10 and Cifar100 datasets. Even compared with the full-precision models of Cliquenet and ResNet101, our model still achieves the best performance. Furthermore, our model contains 5.2M parameters which is far more lower than the best binary model in [14]. Besides, the bracket shows the accuracy of corresponding unbinarized MBCliqueNet. There is a narrow gap between the binarized model and the unbinarized model. Fig. 3 shows that due to the 'warm restart' learning rate schedule, our model can get a 1% to 2% difference between the 64[th] epochs and the 256[th] epochs.

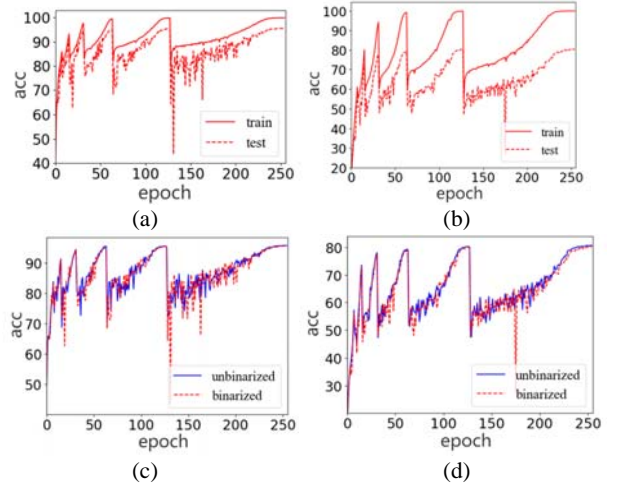

Fig. 3. Training and testing curves of MBCliqueNet and unbinarized MBCliqueNet (u-MBCliquenet). (a) the curve on Cifar10 dataset of MBCliqueNet: (b) the curve on Cifar100 dataset of MBCliqueNet: (c) testing curves on Cifar10 dataset of MBCliquenet and u-MBCliquenet: (d) testing curves on Cifar100 dataset of MBCliquenet and u-MBCliquenet

### 3.5. Results on ImageNet

Due to limitations in computation resources, we only train and test our model on 100-class Tiny ImageNet which are randomly selected from the ImageNet dataset [19] to show the effectiveness of the proposed MBCliquenet. For the experiment we train a MBCliquent with 6 layers in 4 blocks and width of 64-128-128-160. We change the weight decay from 0.0005 to 0.0001. At last we achieve the accuracy of 83.07% and 96.51% for top-1 and top-5, respectively.

## 4. CONCLUSION

In this paper, we propose a new compact and portable deep learning network named MBCliqueNet, which is implemented by a set of binary filters and the proposed modulated operation. Both the binary filters and the M-filters used in modulated operations can be optimized with backpropagation algorithm. As a result, the proposed MBCliqueNet can reduce the required storage space of convolutional filters by a factor of at least 32, in contrast to the full-precision model and achieve better performance than other state-of-art binarized models. The more important point is that even compared with some full-precision models like Resnet, our model preforms better on the dataset we used.

## ACKNOWLEDGEMENT

This work was supported in part by the National Natural Science Foundation of China under Grants 61876037, 31800825, 61871117, 61871124, 61773117, 31571001, 61572258, and in part by the National Key Research and Development Program of China under Grant 2017YFC0107900, and in part by the Short-Term Recruitment Program of Foreign Experts under Grant WQ20163200398.